\documentclass[conference,a4paper]{IEEEtran}
\IEEEoverridecommandlockouts

\usepackage[hidelinks]{hyperref}
\usepackage[cmex10]{amsmath}
\usepackage{amssymb,amsfonts}
\usepackage{dblfloatfix}

\usepackage[ruled,vlined]{algorithm2e}
\usepackage{graphicx}
\graphicspath{{Figures/PDF/}{Figures/PNG/}}
\usepackage{multirow}
\usepackage{booktabs}
\usepackage{siunitx}
\usepackage[numbers,compress]{natbib}
\usepackage{texnames}
\usepackage{bm,bbm}
\usepackage{orcidlink}
\usepackage{algorithmic}
\usepackage[normalem]{ulem}
\useunder{\uline}{\ul}{}
\usepackage[table,xcdraw]{xcolor}

\makeatletter
\DeclareRobustCommand\onedot{\futurelet\@let@token\@onedot}
\def\@onedot{\ifx\@let@token.\else.\null\fi\xspace}

\makeatother

\usepackage[capitalize]{cleveref}
\crefname{section}{Sec.}{Secs.}
\crefname{table}{Tab.}{Tabs.}
\Crefname{section}{Section}{Sections}
\Crefname{table}{Table}{Tables}

\def\Jacksonville{\textbf{Jacksonville}}
\def\Omaha{\textbf{Omaha}}

\usepackage{amsfonts}          
\newcommand{\ind}{\mathbb{1}}  
\begin{document}

\title{Geometric Consistency Protocol for Foundation Model Features in Multi-View Satellite Imagery}

\author{
    \IEEEauthorblockN{
        Qiyan Luo$^*$, Jie Yang$^*$, Yingdong Pi\textsuperscript{\textdagger}, Lekang Wen, Mi Wang
    }
    \IEEEauthorblockA{
        \textit{State Key Laboratory of Information 
Engineering in Surveying, Mapping 
and Remote Sensing, Wuhan University}\\
        430079 Hubei, China\\
        \{luoqy26, yangjie1, pyd\_imars, fanzhongli, wenlk3, wangmi\}@whu.edu.cn
    }
    \thanks{\textsuperscript{*}These authors contributed equally to this work.}
    \thanks{\textsuperscript{\textdagger}Corresponding author. }
    \thanks{Supported by the Key Research and Development Program of Hubei Province (2025BEB069) and LIESMARS Special Research Funding and the National Science Fund for Distinguished Young Scholars grant number 62425102.}
}

\maketitle
\begin{abstract}


Standardized evaluation protocols are indispensable for robust benchmarking in remote sensing, particularly as foundation features are increasingly transferred across diverse sensors and complex imaging geometries. In satellite multi-view reconstruction, conventional evaluations relying on unconstrained 2D global matching are often misleading. The Rational Function Model (RFM) and its Rational Polynomial Coefficients (RPC) dictate a curved, height-dependent epipolar geometry that render flat 2D search spaces physically inconsistent. 

We propose a geometry-faithful and reproducible protocol tailored for the RPC framework. Our approach integrates an RPC-projected 3D consistency metric with a geometry-constrained dense matching proxy, specifically evaluating whether similarity responses remain localized and unique under physically plausible search manifolds. A pivotal finding of our joint reporting strategy is the decoupling of semantic agreement and geometric localization: high cross-view similarity at a projected 3D point does not guarantee reliable matchability in practical inference. Our benchmark demonstrates that incorporating geometric constraints is fundamental to the problem definition in satellite imagery. Furthermore, we show that state-of-the-art 2D backbones remain remarkably competitive against specialized 3D-aware models when subjected to this RPC-consistent evaluation.
\end{abstract}


\section{Introduction}
\subsection{Motivation}
The remote sensing community increasingly relies on shared datasets and benchmarks to compare algorithms. However, comparable performance claims also require standardized evaluation protocols and metric definitions that are robust to differences in platforms, implementations, and imaging conditions. This paper targets that need by proposing a geometry-faithful protocol for evaluating multi-view feature extractors in satellite imagery.

Satellite multi-view 3D reconstruction relies on repeatable cross-view correspondence under a Rational Function Model (RFM) camera with Rational Polynomial Coefficients (RPC) \cite{GAO2023446,10974999}. With the rise of visual foundation models, it is now common to reuse pretrained features from DINOv2 \cite{oquabDINOv2LearningRobust2024}, CLIP \cite{pmlr-v139-radford21a}, and SAM \cite{Kirillov_2023_ICCV}. Recent research further proposes multi-view conditioning and 3D supervision to produce features that are consistent across views and scenes, including multi-view foundation models (MVFM) \cite{segreMultiViewFoundationModels2025a} and FiT3D \cite{yueImproving2DFeature2025}.

A common evaluation pattern in multi-view feature papers is sparse matching with an unconstrained 2D global search: for each query, select the maximum-similarity location in the target image or patch and measure the pixel distance to the ground-truth correspondence. In satellite imagery, this can be misleading for at least three reasons. First, RPC epipolar geometry is curved and height dependent, so the feasible search region is intrinsically a 3D decision rather than a flat 2D scanline. Second, satellite images cover large areas and contain repeated structures such as roads, rooftops, and parking lots, making unconstrained argmax matching prone to spurious maxima. Third, satellite images exhibit stronger radiometric variation and regional heterogeneity than common natural-image benchmarks, which changes the ambiguity of similarity responses.

These observations motivate a protocol design. RPC projected 3D feature consistency evaluates feature agreement for the same 3D point after RPC projection, while Dense matching proxy evaluates whether the same feature field yields a sharp and localized response under geometry-consistent search. The two are complementary rather than redundant: high agreement does not automatically imply high matchability.

We focus on two research questions in the satellite RPC setting. We ask whether multi-view features remain consistent for the same 3D point under RPC projection with ground truth height, and whether they yield localized similarity peaks under geometry constrained search that approximates dense matching.

\begin{figure*}[t]
    \centering
    \includegraphics[width=\textwidth]{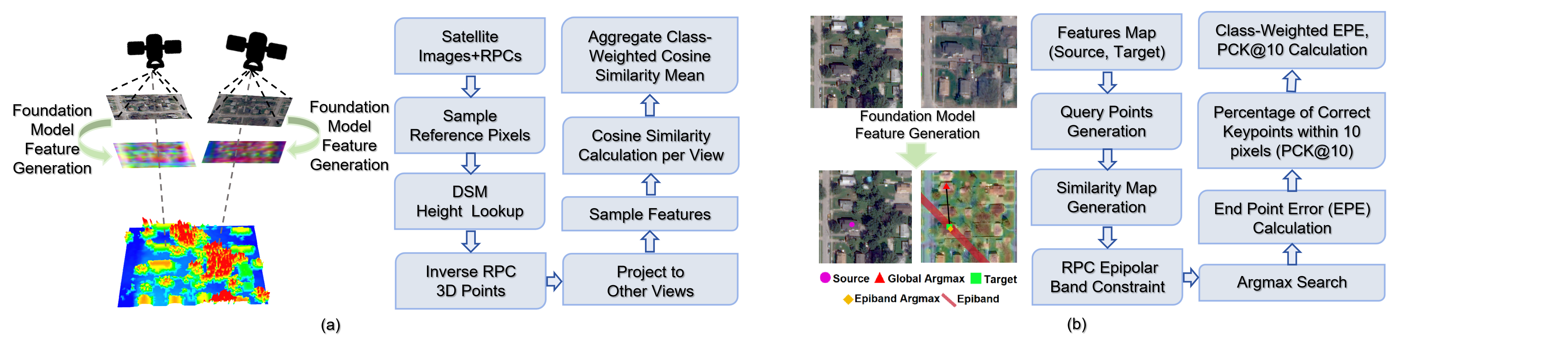}
    \caption{Overview and visualization of the proposed evaluation protocol and metrics. (a) Class-weighted Cosine Similarity, (b) Class-weighted End Point Error and Percentage of Correct Keypoints within 10 pixels.}
    \label{fig:metric1_flow}
\end{figure*}

\subsection{Contributions}
This work proposes a geometric consistency protocol for foundation model features in multi-view satellite imagery. We introduce an RPC projected 3D consistency metric that measures class-weighted cosine similarity across views for the same 3D point, a dense matching proxy with geometry aware inference that reports class-weighted end pointerror and percentage of correct keypoints within 10 pixels under full search and RPC epipolar band constrained search, and a comprehensive evaluation of prevalent foundation model features across more than 100 region tests in Omaha and Jacksonville.

\section{Related Work}
\textbf{2D Visual Foundation Models.} Visual foundation models provide strong generic features for correspondence and reconstruction. DINOv2 \cite{oquabDINOv2LearningRobust2024} improves patch level discrimination and local similarity through joint image and patch objectives. CLIP \cite{pmlr-v139-radford21a} learns an image text embedding that supports semantic alignment and promptable concepts. SAM \cite{Kirillov_2023_ICCV} produces boundary aware representations that are robust to appearance changes. These models are widely reused as frozen feature backbones in downstream pipelines.

\textbf{3D aware and multi-view aware features.} A line of work aims to lift 2D features into 3D consistent representations. Per scene optimization methods fit a 3D representation for each scene, including NeRF \cite{mildenhallNeRFRepresentingScenes2022}, 3D Gaussian Splatting \cite{kerbl3DGaussianSplatting2023}, and grid based fields such as Plenoxels \cite{fridovich-keilPlenoxelsRadianceFields2022} and DFA3D \cite{liDFA3D3DDeformable2023}. These approaches can yield strong geometric consistency but scale poorly when each scene requires its own optimization. Feed forward lifting and multi view conditioning reduce test time optimization. Lift3D \cite{Lift3DZeroShotLifting} uses pretrained geometric proxies for zero shot lifting. FiT3D \cite{yueImproving2DFeature2025} fine tunes 2D backbones with 3D aware supervision distilled from consistent renderings. MVFM \cite{segreMultiViewFoundationModels2025a} conditions features on camera information using ray or Pl\"ucker cues to improve cross view consistency and matching on standard benchmarks. Complementary evaluations analyze reconstruction transformers under sparse overlap and low texture \cite{wuEvaluationDUSt3RMASt3R2025}.

\textbf{Evaluation protocols and reproducibility in remote sensing.} Remote sensing increasingly relies on shared benchmarks and leaderboards \cite{Adorni_2025_CVPR,an2025choicebenchmarkingremotesensing,10641727,11184119}. Reliable comparisons also require consistent split definitions, preprocessing, and metric implementations \cite{10778974}, because small protocol differences can change rankings across teams and platforms. This sensitivity is amplified under cross region and cross constellation transfer \cite{lyu2025deeplearningbaseddomain,tamazyan2025geocrossbenchcrossbandgeneralizationremote}. 

\section{Protocol: Geometry Faithful Evaluation}
\subsection{Data and reporting setting}
We evaluate DFC2019 \cite{c6tm-vw12-19} satellite multi-view imagery with RFM cameras and ground-truth DSM on two full-region test settings with more than 100 areas of interest: Jacksonville and Omaha.

\paragraph{Shared class policy.}
All class-weighted metrics in this paper use one reporting policy shared by Metric~1 and Metric~2. The reporting policy assigns weights of 0.7 and 0.3 to building and ground categories, respectively, to focus on stable geometric surfaces. In the paper, we describe this policy semantically as emphasizing stable and geometrically informative surfaces, especially ground and building categories, while reducing the influence of unstable or weakly informative labels in the headline averages. The normalized weights $\tilde{w}_c$ are always renormalized over the classes present in the evaluated split.

\paragraph{Implementation summary.}
The dataset configuration used by the evaluation code adopts $512\times512$ inputs, $num\_views=4$, crop-based processing, and deterministic DSM-driven sampling. For Metric~1, the released configuration evaluates up to $128$ 3D points per tile and only keeps points visible in at least $3$ views. For Metric~2, the matching script evaluates $512$ correspondences per sample with pair\_mode=all. Correspondence generation and 3D projection rely on the same DSM-constrained fixed-point solver used throughout the repository, with $8$ iterations, a $0.5$\,m stopping threshold, and a symmetric DSM consistency check with a $2$\,m tolerance. When class-stratified reporting is enabled, the scripts scan up to $2048$ candidates per sample to approximate the target class mixture. Invalid points, failed DSM lookups, insufficient-visibility samples, and out-of-bounds reprojections are discarded rather than forced into the metric.

\subsection{Metric 1: RPC projected 3D feature consistency}
We measure whether features are consistent across views for the same 3D point, without relying on a potentially brittle 2D global argmax.

\paragraph{RPC projected correspondence in 3D}
For each sampled reference pixel $\mathbf{x}_i$ in patch coordinates, we obtain a ground-truth height $h_i$ from the DSM and compute a 3D point $\mathbf{X}_i$ by inverse RPC projection in the reference view:
\begin{equation}
    \mathbf{X}_i = \pi^{-1}_{r}(\mathbf{x}_i, h_i).
\end{equation}

In the implementation, the ground point is solved by the DSM-constrained fixed-point RPC procedure described above, using the DSM height at the backprojected location as initialization when available and otherwise falling back to the RPC height offset. We then forward-project $\mathbf{X}_i$ into each other view $j$ using its RPC to obtain the corresponding pixel $\mathbf{x}_{i}^{(j)} = \pi_{j}(\mathbf{X}_i)$ and bilinearly sample the feature map at that sub-pixel location. Points are only counted when the projected correspondence remains valid in enough views to satisfy the minimum-view criterion.

\paragraph{Cosine similarity metrics.}
Let $\mathbf{f}_r(\mathbf{x}_i)$ be the reference feature and $\mathbf{f}_j(\mathbf{x}_{i}^{(j)})$ be the feature in view $j$. With $\tilde{\mathbf{f}} = \mathbf{f}/\lVert\mathbf{f}\rVert_2$, the per-view similarity is
\begin{equation}
    s_{i,j} = \left\langle \tilde{\mathbf{f}}_r(\mathbf{x}_i), \tilde{\mathbf{f}}_j(\mathbf{x}_{i}^{(j)}) \right\rangle.
\end{equation}

We average over the set of valid projected views $\mathcal{V}_i$ to obtain per-point similarity $s_i$:
\begin{equation}
    s_i = \frac{1}{|\mathcal{V}_i|} \sum_{j \in \mathcal{V}_i} s_{i,j}.
\end{equation}

Let $N$ be the number of valid sampled points. Let $y_i \in \mathcal{C}$ be the land-cover label for point $i$, $N_c = |\{i : y_i = c\}|$, and let $\tilde{w}_c$ be the normalized class weights defined above. We report mean cosine similarity $\frac{1}{N}\sum_i s_i$ and the class-weighted mean cosine similarity
\begin{equation}
    \text{clsCos} = \sum_{c \in \mathcal{C}} \tilde{w}_c \cdot \frac{1}{N_c} \sum_{i:y_i=c} s_i.
\end{equation}

\begin{figure}
    \centering
    \includegraphics[width=1.00\linewidth]{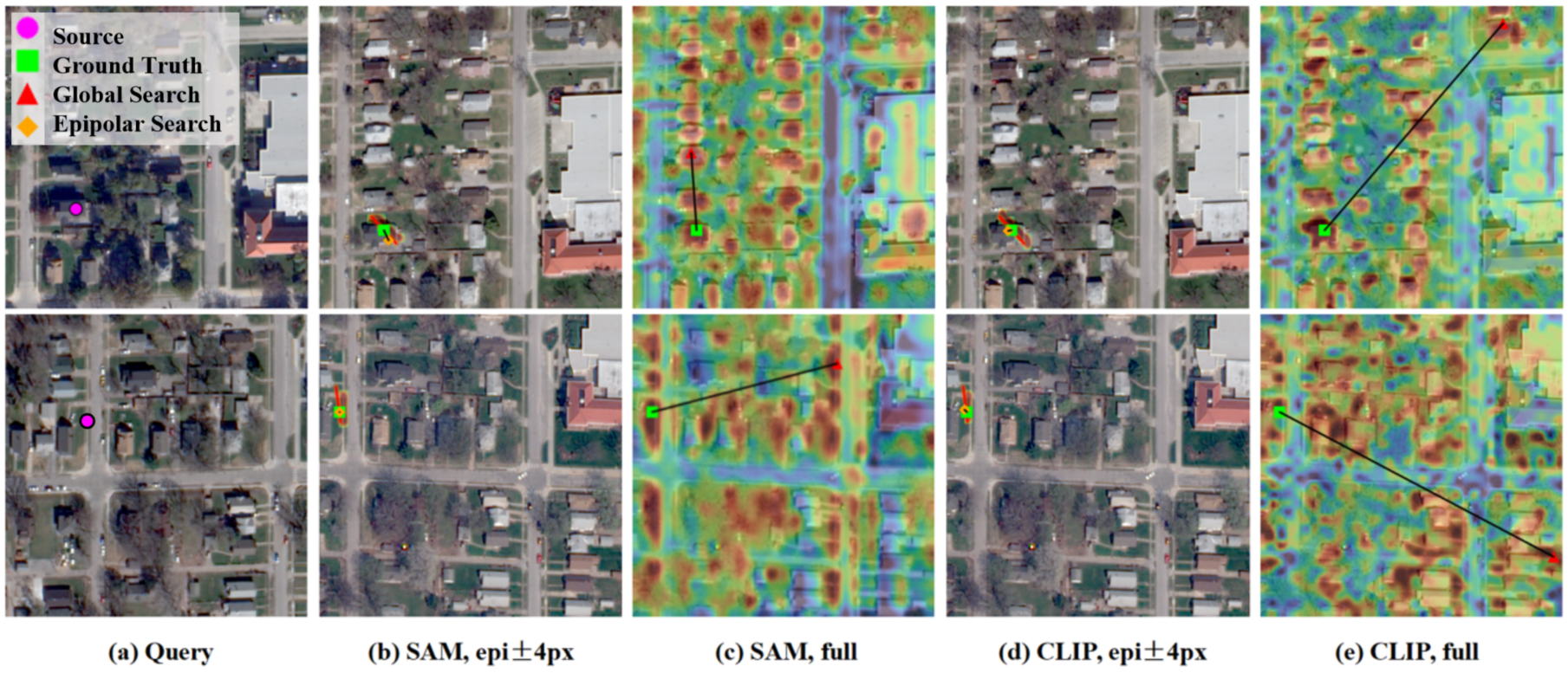}
    \caption{Qualitative example of similarity responses and geometry constrained matching.}
    \label{fig:placeholder}
\end{figure}

\begin{table}[t]
    \centering
    \scriptsize
    \caption{RPC projected 3D consistency clsCos.}
    \label{tab:3d_cosine}
    \begin{tabular}{lcc}
        \toprule
        \multirow{2}{*}{Method} & \multicolumn{2}{c}{clsCos $\uparrow$} \\
        \cmidrule(lr){2-3}
        & Omaha & Jacksonville \\
        \midrule
        DINOv2-Small \cite{oquabDINOv2LearningRobust2024} & 0.847 & 0.875 \\
        DINOv2-Small reg \cite{oquabDINOv2LearningRobust2024} & 0.872 & 0.907 \\
        DINOv3-Small \cite{simeoni2025dinov3}& 0.898 & 0.907 \\
        SAM-Base \cite{Kirillov_2023_ICCV} & 0.866 & 0.846 \\
        CLIP-Base \cite{pmlr-v139-radford21a}& 0.563 & 0.630 \\
        FiT3D (CLIP-Base) \cite{yueImproving2DFeature2025} & 0.925 & 0.947 \\
        FiT3D (DINOv2-Small reg) \cite{yueImproving2DFeature2025} & 0.905 & 0.927 \\
        FiT3D (DINOv2-Small) \cite{yueImproving2DFeature2025} & 0.944 & 0.953 \\
        MVFM (CLIP-Base) \cite{segreMultiViewFoundationModels2025a} & 0.785 & 0.804 \\
        MVFM (DINOv3-Small) \cite{segreMultiViewFoundationModels2025a}& 0.865 & 0.882 \\
        MVFM (DINOv2-Small reg) \cite{segreMultiViewFoundationModels2025a} & \textbf{0.999} & \textbf{0.999} \\
        MVFM (DINOv2-Small reg, w/o PL) \cite{segreMultiViewFoundationModels2025a} & \textbf{0.999} & \textbf{0.999} \\
        MVFM (SAM-Base) \cite{segreMultiViewFoundationModels2025a} & 0.979 & 0.978 \\
        \bottomrule
    \end{tabular}
\end{table}

\subsection{Metric 2: Dense matching proxy}
Given a set of query pixels in a reference view and ground-truth correspondences in a target view, we predict correspondences using feature similarity. We report End Point Error (EPE) and Percentage of Correct Keypoints within 10 pixels (PCK@10), and we further report class-weighted variants (clsEPE and clsPCK@10) to emphasize the same task-relevant classes used in Metric~1.

\subsection{Dense matching proxy: global search vs epipolar band search}
\begin{table*}[ht]
    \centering
    \scriptsize
    \caption{Dense matching proxy clsEPE and clsPCK@10 on \Omaha and \Jacksonville.}
    \label{tab:matching_joint}
    \begin{tabular}{lcccccccc}
        \toprule
        Method & \multicolumn{4}{c}{Omaha} & \multicolumn{4}{c}{Jacksonville} \\
        \cmidrule(r){2-5} \cmidrule(l){6-9}
        & \multicolumn{2}{c}{Epipolar Band Search (±4px)} & \multicolumn{2}{c}{Global Search} & \multicolumn{2}{c}{Epipolar Band Search (±4px)} & \multicolumn{2}{c}{Global Search} \\
        \cmidrule(r){2-3} \cmidrule(r){4-5} \cmidrule(l){6-7} \cmidrule(l){8-9}
        & clsEPE $\downarrow$ & clsPCK@10 $\uparrow$ & clsEPE $\downarrow$ & clsPCK@10 $\uparrow$ & clsEPE $\downarrow$ & clsPCK@10 $\uparrow$ & clsEPE $\downarrow$ & clsPCK@10 $\uparrow$ \\
        \midrule
        CLIP-Base \cite{pmlr-v139-radford21a}& 15.174 & 0.405 & 118.788 & 0.062 & 15.723 & 0.380 & 130.403 & 0.045 \\
        DINOv2-Small  \cite{oquabDINOv2LearningRobust2024}& 13.277 & 0.499 & 74.392 & 0.131 & 13.593 & 0.476 & 83.318 & 0.108 \\
        DINOv2-Small reg \cite{oquabDINOv2LearningRobust2024}& 13.495 & 0.490 & 83.809 & 0.116 & 13.809 & 0.469 & 96.635 & 0.095 \\
        DINOv3-Small \cite{simeoni2025dinov3}& \textbf{12.103} & 0.524 & \textbf{37.526} & \textbf{0.294} & \textbf{12.114} & 0.513 & \textbf{24.352} & \textbf{0.295} \\
        MVFM (DINOv3-Small) \cite{yueImproving2DFeature2025} & 13.590 & 0.464 & 55.978 & 0.174 & 13.586 & 0.451 & 39.481 & 0.172 \\
        SAM-Base \cite{Kirillov_2023_ICCV} & 12.186 & \textbf{0.532} & 83.515 & 0.212 & 12.624 & \textbf{0.525} & 86.603 & 0.234 \\
        FiT3D (CLIP-Base) \cite{yueImproving2DFeature2025} & 15.368 & 0.387 & 114.442 & 0.060 & 14.997 & 0.412 & 103.532 & 0.076 \\
        FiT3D (DINOv2-Small reg) \cite{yueImproving2DFeature2025} & 13.725 & 0.473 & 93.126 & 0.105 & 13.333 & 0.501 & 83.370 & 0.127 \\
        FiT3D (DINOv2-Small) \cite{yueImproving2DFeature2025} & 14.457 & 0.440 & 99.455 & 0.072 & 14.267 & 0.458 & 104.810 & 0.078 \\
        MVFM (CLIP-Base) \cite{segreMultiViewFoundationModels2025a} & 15.723 & 0.380 & 130.403 & 0.045 & 15.174 & 0.405 & 118.788 & 0.062 \\
        MVFM (DINOv2-Small reg) \cite{segreMultiViewFoundationModels2025a} & 16.172 & 0.396 & 69.811 & 0.044 & 16.017 & 0.413 & 84.505 & 0.045 \\
        MVFM (DINOv2-Small reg, w/o PL) \cite{segreMultiViewFoundationModels2025a} & 15.811 & 0.410 & 69.474 & 0.050 & 15.657 & 0.426 & 84.590 & 0.050 \\
        MVFM (SAM-Base) \cite{segreMultiViewFoundationModels2025a} & 12.611 & 0.522 & 95.056 & 0.172 & 13.278 & 0.505 & 98.729 & 0.183 \\
        \bottomrule
    \end{tabular}
\end{table*}

\paragraph{Metrics.}
Let $N$ be the number of evaluated correspondences. Let the predicted match be $\hat{\mathbf{p}}_i = (\hat{u}_i, \hat{v}_i)$ and the ground truth be $\mathbf{p}_i = (u_i, v_i)$ in target-image pixels. The Euclidean error is $e_i = \lVert \hat{\mathbf{p}}_i - \mathbf{p}_i \rVert_2$. We define
\begin{equation}
    \text{EPE} = \frac{1}{N} \sum_{i=1}^{N} \left\lVert \hat{\mathbf{p}}_i - \mathbf{p}_i \right\rVert_2,
\end{equation}
\begin{equation}
    \text{PCK@10} = \frac{1}{N} \sum_{i=1}^{N} \ind\left[\left\lVert \hat{\mathbf{p}}_i - \mathbf{p}_i \right\rVert_2 < 10\right].
\end{equation}
With the same class weights $\tilde{w}_c$ as above, we define
\begin{equation}
    \text{clsEPE} = \sum_{c \in \mathcal{C}} \tilde{w}_c \cdot \frac{1}{N_c} \sum_{i : y_i = c} e_i,
\end{equation}
\begin{equation}
    \text{clsPCK@10} = \sum_{c \in \mathcal{C}} \tilde{w}_c \cdot \frac{1}{N_c} \sum_{i : y_i = c} \ind[e_i < 10].
\end{equation}

\paragraph{Inference modes}
We report both unconstrained global search over the target patch and RPC epipolar band search that restricts candidates to a band around the curved and height dependent epipolar curve.

\section{Methods Under Test}
We treat all methods as frozen feature extractors under the same RPC consistent protocol and group them by how geometry enters feature learning. As general purpose 2D backbones, DINOv2 \cite{oquabDINOv2LearningRobust2024} provides strong self-supervised dense features, DINOv3 \cite{simeoni2025dinov3} scales curated data and training stabilization to improve dense feature quality without fine tuning, CLIP \cite{pmlr-v139-radford21a} learns transferable representations from large scale image text alignment for open vocabulary semantics, and SAM \cite{Kirillov_2023_ICCV} provides promptable segmentation features trained on a large mask dataset that generalize across distributions. Building on such 2D features, FiT3D \cite{yueImproving2DFeature2025} performs 3D aware fine tuning to encourage multi-view consistency, while MVFM \cite{segreMultiViewFoundationModels2025a} uses camera conditioned multi-view adapters with ray or Pl\"ucker cues to promote cross view feature alignment, enabling us to test whether improving semantic consistency also improves geometric matchability in satellite imagery.

\section{Results and Analysis}

\textbf{Interpretation of Metrics.}
We distinguish between Metric~1 (class-weighted cosine similarity, clsCos), which measures feature agreement at the projected 3D point, and Metric~2 (class-weighted end-point error/ class-weighted percentage of correct keypoints, clsEPE/clsPCK), which evaluates matching localization. \Cref{tab:3d_cosine} shows that while 3D-aware models (FiT3D, MVFM) significantly boost feature consistency, this does not always translate to better matching. This decoupling occurs because high consistency can coexist with broad similarity plateaus, whereas satellite reconstruction demands sharp, unique peaks for geometric discriminability.

\textbf{The Necessity of Epipolar Constraints.}
A key observation is that unconstrained global matching is an ill-posed evaluation for satellite imagery. As shown in \Cref{tab:matching_joint}, all methods exhibit a dramatic performance jump when moving from global search to epipolar-band search (±4 pixels). This confirms that geometry-constrained inference is essential to filter out physically implausible matches caused by repeated structures (e.g., rooftops, roads) and long baselines.

\textbf{Methodology Comparison.}
Under the RPC-consistent protocol, strong 2D backbones remain highly competitive. \textbf{DINOv3-Small} demonstrates superior discriminative localization in global search, while \textbf{SAM} shows the highest stability under epipolar constraints. Multi-view-aware models, though improved in semantic alignment, do not consistently outperform these 2D baselines in precise geometric matching, highlighting a trade-off between feature invariance and peak sharpness.

Natural-image protocols often assume calibrated or near-pinhole cameras where rectification yields near-straight epipolar lines, making correspondence evaluation close to a 1D search. Satellite imagery violates these assumptions. RPC epipolar geometry is curved and height dependent, so feasible correspondences form a 3D manifold rather than a scanline. Satellite images also cover larger extents with frequent self-similarity, and multi-temporal or cross-sensor settings introduce radiometric shifts and real change, all of which reshape similarity responses. As a result, unconstrained global argmax matching is ill-posed and can change method rankings by permitting physically implausible matches.

\textbf{Key observations.}
First, epipolar-band-constrained inference is essential: replacing global search with \texttt{epi-4} improves clsPCK@10 for every method. Second, strong 2D backbones remain competitive. SAM is the strongest and most stable under epipolar-band search, while DINOv3-Small is especially strong under global search. Third, multi-view and 3D-aware families are competitive but do not consistently outperform the strongest 2D baselines under the same RPC-consistent inference. 

\section{Conclusion}
This paper presents a geometry-faithful and reproducible protocol for evaluating satellite multi-view feature extractors under the RPC camera model. The protocol combines RPC-projected 3D feature consistency with geometry-constrained dense matching metrics, and it makes explicit that these two properties are complementary rather than interchangeable. Experiments on full-region tests in Omaha and Jacksonville show that geometry-constrained inference is necessary for meaningful correspondence evaluation and that semantic consistency does not necessarily imply geometric matchability. These findings support the use of unified protocol choices and metric definitions when comparing emerging foundation features in remote sensing.

\section{Limitations and Future Work}
Our findings advocate for the adoption of unified, geometry-aware evaluation standards to accurately benchmark emerging foundation models in the remote sensing domain. While this study establishes a reproducible protocol, it is limited to two DFC2019 regions and a selection of foundation models. 

Future extensions will include: (1) expanding to cross-sensor and multi-temporal imagery; (2) conducting sensitivity analyses on epipolar-band width and DSM noise; and (3) releasing a standardized evaluation toolkit to facilitate cross-institutional benchmarking.

\small
\bibliographystyle{IEEEtranN}
\bibliography{ref}

\end{document}